\documentclass[conference]{IEEEtran}
\IEEEoverridecommandlockouts
\usepackage{amsmath,amsfonts}
\usepackage{algorithmic}
\usepackage{algorithm}
\usepackage{array}
\usepackage[caption=false,font=normalsize,labelfont=sf,textfont=sf]{subfig}
\usepackage{textcomp}
\usepackage{stfloats}
\usepackage{mathtools}
\usepackage[table]{xcolor}
\usepackage{color, colortbl}
\usepackage{balance}
\usepackage{url}
\definecolor{LightCyan}{rgb}{0.88,1,1}
\definecolor{mintcream}{rgb}{0.96, 1.0, 0.98}
\usepackage{tabularx,booktabs}
\usepackage{adjustbox}
\usepackage{mathrsfs}
\usepackage{textcomp}
\usepackage{hyperref}
\usepackage{enumitem}
\usepackage{amsmath}
\usepackage{enumitem}
\usepackage{verbatim}
\usepackage{graphicx}
\usepackage{cite}
\usepackage{lipsum}

\begin{document}

\title{Personalized Federated Learning for Statistical Heterogeneity\\
}


\author{\IEEEauthorblockN{1\textsuperscript{st}Muhammad Firdaus}
\IEEEauthorblockA{\textit{Department of AI Convergence} \\
\textit{Pukyong National University}\\
Busan, Republic of Korea \\
0000-0003-0104-848X}
\and
\IEEEauthorblockN{2\textsuperscript{nd} Kyung-Hyune Rhee}
\IEEEauthorblockA{\textit{Division of Computer Engineering and AI} \\
\textit{Pukyong National University}\\
Busan, Republic of Korea \\
0000-0003-0466-8254}
}

\maketitle

\begin{abstract}

The popularity of federated learning (FL) is on the rise, along with growing concerns about data privacy in artificial intelligence applications. FL facilitates collaborative multi-party model learning while simultaneously ensuring the preservation of data confidentiality. Nevertheless, the problem of statistical heterogeneity caused by the presence of diverse client data distributions gives rise to certain challenges, such as inadequate personalization and slow convergence. In order to address the above issues, this paper offers a brief summary of the current research progress in the field of personalized federated learning (PFL). It outlines the PFL concept, examines related techniques, and highlights current endeavors. Furthermore, this paper also discusses potential further research and obstacles associated with PFL.

\end{abstract}


\section{Introduction}

Recently, we have observed favorable advancements through the utilization of machine learning (ML) support. ML facilitates model training by consolidating and aggregating client data on a central server. However, these techniques encounter significant privacy concerns, including the potential for sensitive data exposure, the risk of single points of failure (SPoF), and the substantial resource burdens associated with collecting and storing training data. Moreover, due to the introduction of data privacy regulations like the Health Insurance Portability and Accountability Act (HIPAA) \cite{ref1} and the General Data Protection Regulation (GDPR) \cite{ref2}, there is a growing demand for safeguarding privacy in AI, as highlighted by Cheng et al. \cite{ref3}.
To tackle these challenges, the concept of federated learning (FL) emerged as a promising approach. FL allows distributed mobile devices to collectively train models without centralizing the training data, thus maintaining local data on the client's devices. Furthermore, FL showcases its effectiveness and privacy preservation by enabling collaborative local training and shared updates to the machine learning model, all while safeguarding the confidentiality of individual datasets \cite{ref4}.


However, despite their advantages, the existing methods of FL encounter several difficulties. These challenges include statistical heterogeneity, resulting in poor convergence and a lack of personalization due to the influence of non-independent and identically distributed (non-IID) data distribution across clients. As a result, these challenges have a detrimental impact on the overall performance of the global FL model when applied to individual clients with eminent performance outputs. Consequently, some clients are reluctant or even unwilling to participate in and contribute to the collaborative FL training. In order to address these issues, the concept of personalized federated learning (PFL) has been introduced \cite{ref5}.
This paper provides an overview of the current research landscape in PFL, highlighting its high-level aspects. Additionally, the paper briefly outlines the core concept of PFL and delves into the associated methodologies and ongoing research. Lastly, we also discuss observations and challenges in the field of PFL.


The paper's structure is organized in the following manner: Section 2 provides an explanation of the background knowledge pertaining to FL and the impact of non-IID data. Section 3 provides a concise overview of the concept of PFL. Subsequently, Section 4 provides an overview of the survey conducted on pertinent methodologies and ongoing research efforts pertaining to PFL. In Section 5, we examine several observations and concerns. Finally, this work is concluded in Section 6. 

\section{Preliminaries}
\label{sect:preliminaries}

\subsection{Federated Averaging (FedAvg)}
\label{sect:fedAvg}

In the context of the client-server architecture in traditional ML, the process of training models is consistently executed on the server \cite{ref6}. Clients solely perform as data providers, whereas the server also accomplishes data training and aggregation. However, several concerns are associated with the classical ML strategy, especially regarding user privacy. Therefore, Google introduced federated averaging (FedAvg) as a new communication-efficient optimization algorithm for FL. FedAvg \cite{ref7} is an effective approach for training models in a distributed manner. This method enables mobile devices to collaborate in training the models without the need to centralize the training data, hence allowing the local data to be stored on the respective mobile devices. 

\begin{figure}[ht]
	\begin{centering}
	\includegraphics[width=0.5 \textwidth]{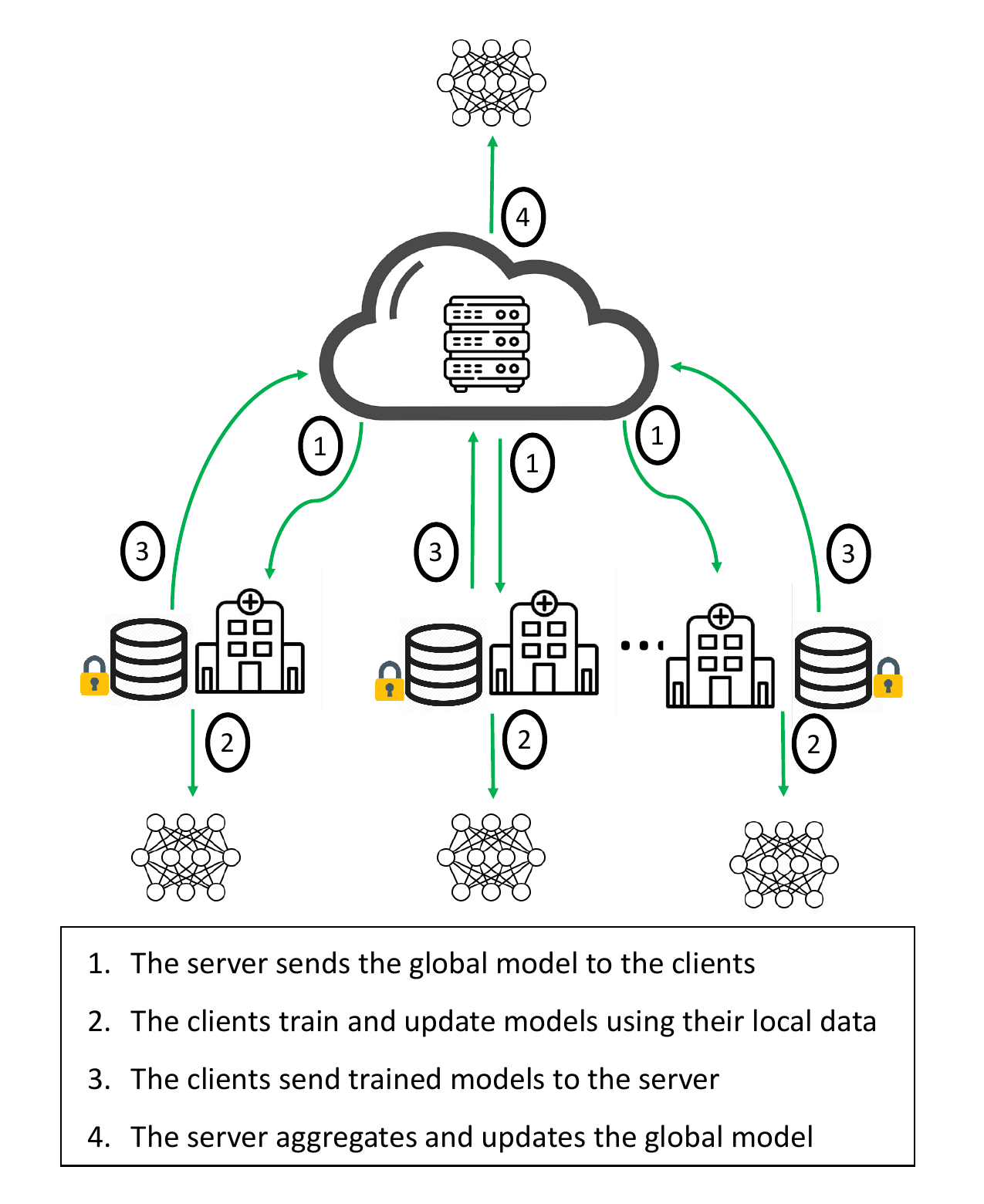}
	\caption{The illustration of FedAvg procedures}
	\label{fig:fedavg}
	\end{centering}
\end{figure}


The illustration of the FedAvg processes is depicted in Figure \ref{fig:fedavg}. Initially, the server, acting as a provider of models, transmits the global model to the clients. Participating clients obtain the global model from the aggregator server, train the current global model on their local data, and then upload the resulting model to the aggregator server. In its role as an aggregator server, the central server collects and consolidates all the model updates from the clients in order to generate an updated global model for the following iteration. Therefore, the FedAvg algorithm efficiently enhances the privacy of clients by preventing some attacks that aim to get unauthorized access to the local training data \cite{ref8}. Additionally, it demonstrates improved communication efficiency compared to traditional distributed stochastic gradient descent (SGD) methods by reducing the number of communication rounds. Moreover, the efficacy of FedAvg is evident in various domains of learning, including predictive models in health \cite{ref9}, low latency vehicle-to-vehicle communication \cite{ref10}, vocabulary estimation \cite{ref11}, and next word prediction \cite{ref12}. In each of these instances, FedAvg consistently demonstrates enhanced performance.

\subsection{Statistical Heterogeneity: The Effect of Non-IID Data}
\label{sect:blockchain}


While FL offers numerous benefits for various real-world use cases, it also presents several challenges due to the diverse data distribution among clients. This diversity leads to a deficiency in personalization and poor convergence. The data distribution among clients is notably dissimilar, resulting in a situation where the data distribution is mostly non-IID. Consequently, this statistical disparity complicates the training of a single model capable of performing effectively for all clients. Furthermore, the presence of non-IID data can substantially impact the accuracy of the FedAvg algorithm. The local objectives of each client do not align with the global optimum due to the significant differences in data distribution between individual local datasets and the overarching global distribution. As a consequence, a drift occurs in the updated local models\cite{ref13}. In other words, each model is updated during the local training phase towards its specific local optima, which could be far away from the global optima. Primarily, when numerous noteworthy local updates take place (indicating a considerable count of local epochs), the resultant aggregated model may potentially drift from the global optima \cite{ref14},\cite{ref15}. As a result, the convergence of the global model is notably less precise compared to the scenario where data distribution is independent and identical (IID).

\section{Personalized Federated Learning (PFL) Concept}
\label{sect:concept}

The concept of PFL comes from shortcomings created by statistical heterogeneity and the non-IID data distribution that force the need for personalization methods in the current FL setting. The accuracy of FedAvg-based techniques suffers a substantial decrease when training non-IID data, mainly due to client drift. The study conducted by Tan et al. \cite{ref16} presents two distinct approaches for achieving PFL. The approaches encompass the global model personalization strategy, which involves the training of a singular global model, and the learning personalized models strategy, which entails the individual training of PFL models. The majority of personalization techniques for a global model FL \cite{ref17} often involve two separate procedures: the initial stage entails the collaborative development of a global model, followed by the subsequent phase, which entails utilizing the client's private data to personalize the global model \cite{ref18}. In summary, the PFL technique commonly adopts FedAvg as the de facto method for training in broad FL environments. The difference is that after the training process, the trained global model is personalized through additional training and local adaptation steps tailored to the local dataset of the client. The objective of PFL is stated as follows.

\begin{equation}
 \min_{w \in \mathbb{R} ^d}F\left ( w \right ) = \frac{1}{M} \sum_{m=1}^M f_m({w_m}+\frac{\mu}{\alpha_i})
\end{equation}

\noindent where $F_m$ is a loss function associated with the $m-$th clients; $w\in \mathbb{R}^d$ encodes the parameters of local model $w$, $\mu$ is the regularization parameter, and $\alpha_p$ represents the local adaptation steps in iteration $i$.


On the other hand, the learning personalized model strategy aims to apply different learning algorithms and modify the process of FL aggregation to achieve PFL by classifying it into architecture-based and similarity methods. Moreover, the objective of PFL is to collaboratively train personalized models for a particular group of clients by leveraging the non-IID nature of their respective data distribution along with simultaneously protecting their private information \cite{ref19}. Further, the three following goals must all be addressed simultaneously rather than separately in improving a more valuable PFL for practical concerns \cite{ref20}: 
\begin{itemize}
  \item Attaining swift model convergence within a limited set of training iterations.
  \item Enhancing personalized models that are advantageous to a significant portion of clients.
  \item Formulating more precise overarching models that aid clients possessing restricted private data in achieving personalization.
\end{itemize}

\section{Survey of Related Methods and Current Works for PFL}
\label{sect:survey}
In this section, we survey the related methods and current works of PFL by separating them into three classifications (i.e., based on data, model, and similarity) for manageable knowledge. Moreover, we provide a classification diagram of PFL that can be seen in Figure 2. Finally, Table 1 summarizes the state-of-the-art methods for the PFL research area.


\begin{figure*}[t!]
  \includegraphics[width=\textwidth,height=10.8cm]{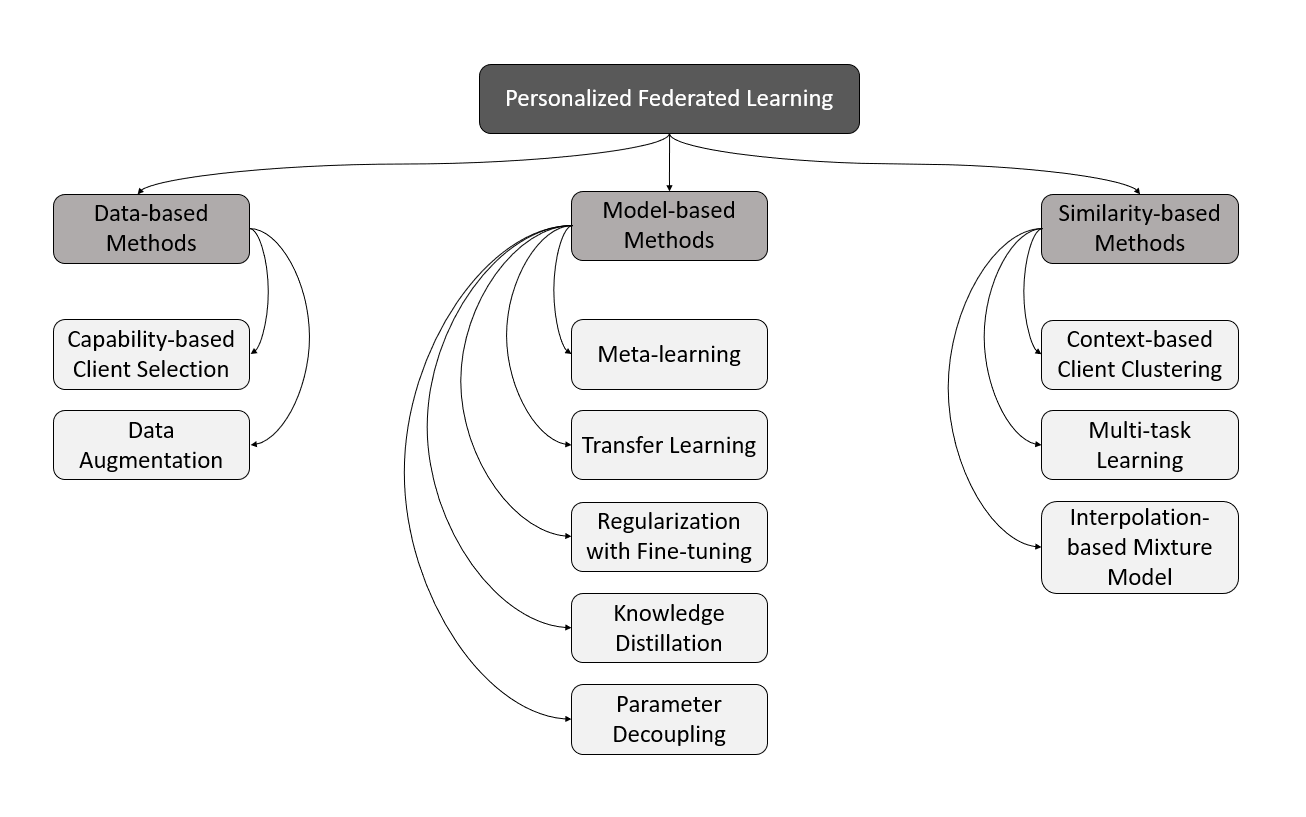}
  \caption{The classification diagram of PFL}
  \label{fig:class}
\end{figure*}



\subsection{Data-based Methods}
Data-based methods desire to diminish the client data distributions' statistical heterogeneity, which is encouraged by the client drift problem. This method consists of capability-based client selection and data augmentation that supports improving the global FL model's generalization performance.

\paragraph{Capability-based Client Selection}: In each FL communication round, these methods optimize the subset of involved clients to enhance the model generalization performance and facilitate sampling from a more uniform distribution of client's data. The authors in \cite{ref21} proposed FAVOR, a client selection algorithm based on deep Q-learning formulation that aims to minimize the number of communication rounds while maximizing accuracy. To mitigate the impact of non-IID data, FAVOR was designed to selectively choose a certain number of clients to participate in each training round. Nevertheless, deep Q-learning demands computationally higher algorithms that lead to increased computation overhead. 

\paragraph{Data Augmentation}: These approaches are simple to implement in the conventional FL setting that is offered to improve data statistical homogeneity and reduce data inequality. In the study conducted by Zhao et al. \cite{ref22}, a data sharing approach was introduced. This technique aims to disseminate a proportionate subset of global data while addressing the issue of imbalanced allocation of client data classes among individual clients. However, potential privacy leakage is the main challenge since the client's data distribution is frequently shared during training.

\subsection{Model-based Methods}
Model-based methods obey the prevailing FL training setting that involves training using a single global model. This method consists of meta-learning, transfer learning, regularization with fine-tuning, knowledge distillation, and parameter decoupling.

\paragraph{Meta-learning}: This method strives to solve new tasks with only a few training examples by generating highly-adaptable models and applying training on multiple learning tasks that can further learn. Moreover, the meta-learning method, also known as initialization-based learning, provides fast adaptation to new tasks with better generalization. Authors in \cite{ref20} highlight the relationship between FedAvg and model agnostic meta-learning (MAML). The MAML algorithm is executed in two distinct phases, namely meta-testing and meta-training. They note that solely optimizing for global model accuracy may not lead to strong personalization results.

\begin{table*}[t!]
  \caption{Summary of the state-of-the-art methods for the PFL research.}
  \label{tab:table}
  \begin{adjustbox}{width=1\textwidth}
  \begin{tabular}{p{0.15\linewidth}  p{0.07\linewidth}  p{0.13\linewidth}  p{0.15\linewidth}  p{0.3\linewidth} p{0.25\linewidth}}
    \toprule
    \textbf{Paper Ref.} & \textbf{Year} & \textbf{PFL Protocols} & \textbf{PFL Methods} & \textbf{Key Contributions} & \textbf{Remarks}\\
    \midrule
    \midrule
    Wang, et al. \cite{ref21}    & 2020     & FAVOR     & Capability-based client selection   & Enhancing the model generalization performance.  &High communication and computation cost\\
  \bottomrule 

   Zhao, et al. \cite{ref22}    & 2018     & Based on FedAvg     & Data augmentation   & These methods are easy to enforce in the standard FL setting.  &Potential privacy leakage is the main challenge\\
  \bottomrule 

  Jiang, et al. \cite{ref20}    & 2019     & MAML on FedAvg     & Meta-learning   &  Highlight the relationship between model agnostic meta-learning (MAML) and FedAvg.  &A single global model setting with centralized approach\\
  \bottomrule 

  Chen, et al. \cite{ref23}    & 2020     & FedHealth     & Transfer learning   &  Reducing the domain dissimilarity between the local models and global model to improve personalization.  &A single global model setting and potential attacks considerations\\
  \bottomrule 

  Li, et al. \cite{ref14}    & 2020     & FedProx     & Regularization with fine-tuning   &  Evaluates the dissimilarity between the global FL model and local models.  &A single global model setting with centralized approach\\
  \bottomrule 

  Li and wang, et al. \cite{ref24}    & 2019     & FedMD     & Knowledge distillation   &  Supports resource heterogeneity for each client with communication-efficient.  &Requires the representative proxy datasets and benchmark\\
  \bottomrule

  Arivazhagan, et al. \cite{ref25}    & 2019     & FedPer     & Parameter decoupling   &  Simple formulation and allows some model layers stored locally while the remaining layers are trained using FL.  &Difficult to determine optimal system and requires the representative proxy datasets and benchmark\\
  \bottomrule

  Sattler, et al. \cite{ref26}    & 2020     & FMTL     & Context-based client clustering   &  Clustering approaches are beneficial when client partitions are inherently present.  &Requires higher computational overhead and potential attacks considerations \\
  \bottomrule

  Smith, et al. \cite{ref27}    & 2017     & MOCHA     & Multi-task learning   &  Utilize client relationships in pairs to discover models that are comparable for similar clients.  &Requires the representative proxy datasets and benchmark \\
  \bottomrule

  Peterson, et al. \cite{ref28}    & 2019     & Not defined     & Interpolation-based Mixture Model   &  leverages a mixture of local models and global model.  &A single global model with centralized approach\\
  \bottomrule

\end{tabular}
\end{adjustbox}
\end{table*}

\paragraph{Transfer Learning}: This method transfers the trained model parameters, called knowledge, from a source to a destination to avoid the necessity to construct models from scratch. Moreover, it lowers the domain dissimilarity between the local and global models to improve personalization. In \cite{ref23}, the authors propose FedHealth, a novel federated transfer learning framework tailored for wearable healthcare, to address these challenges. They transfer the formerly trained global model to each device after first training a global model using standard FL. Additionally, FedHealth leverages FL for data aggregation and employs transfer learning to build relatively personalized models.

\paragraph{Regularization with Fine-tuning}: Model regularization is a standard method to improve convergence and prevent overfitting during model training. In order to tackle client drift, regularization is implemented between the local models and global model. The work in \cite{ref14} proposed FedProx to alter the effect of updated models by considering the disparity between the local models and the global FL model. FedProx is a generalized and re-parametrized version of FedAvg, designed to address heterogeneity in federated networks. 

\paragraph{Knowledge Distillation}: The primary purpose of this method is to support resource heterogeneity for each client with communication-efficient. The authors in\cite{ref24}  proposed FedMD protocol to design a high degree of flexibility for clients through PFL that facilitates creating various models utilizing their private data. This framework accommodates scenarios where each participant designs their own model due to intellectual property concerns and varying tasks and data.

\paragraph{Parameter Decoupling}: This method seeks to gain PFL  from the global FL model parameters by decoupling the local private model parameters. In \cite{ref25}, the authors suggested that some model layers can be stored locally while the remaining layers are trained using FL. In this sense, this approach includes a base model along with a personalization layer to address the challenge of statistical heterogeneity in FL.

\subsection{Similarity-based Methods}
Similarity-based methods seek to accomplish PFL by mapping client associations. Every client learns a personalized model, while the associated clients learn similar models. PFL has investigated many client relationship types, including clustering takes into account group-level client relationships, whereas model interpolation and multi-task learning take into account pairwise client ties.

\paragraph{Context-based Client Clustering}: When client partitions are inherently present, clustering approaches are beneficial. It is more appropriate to use a multi-model method where an FL model is trained for every similar group of clients. Hierarchical clustering has been added to FL in \cite{ref26} as a postprocessing step. The clients in FL are grouped together using an optimal partitioning technique that relies on the cosine similarity of the gradient updates generated by the clients.

\paragraph{Multi-task Learning}: Smith et al. \cite{ref27}  introduced MOCHA, an algorithm designed to address the challenge of federated multitask learning. This method aims to learn both user task settings and a similarity matrix concurrently. The distributed multitask MOCHA algorithm tackles several learning challenges, such as communication restrictions, stragglers, and fault tolerance. They concentrate on the convex case, and it is opaque how they involve deep learning models that are not convex when powerful duality is no longer assured.

\paragraph{Interpolation-based Mixture Model}: This method is a simple formulation that leverages a mixture of global model and local models. In \cite{ref28}, the authors suggested using methods from various expert literature. Their method of model interpolation learns an interpolation weight depending on features to improve model accuracy for all users. They validate their approach using both real and synthetic data, highlighting its effectiveness in enhancing FL performance.

\section{Discussion and Remarks}
\label{sect:discussion}
Recently, PFL research has been garnering attention due to its potential to tackle the fundamental challenges of current FL, including statistical heterogeneity that leads to poor convergence and a lack of personalization because of the influence of non-IID data distribution across clients. However, based on the explanation in Table \ref{tab:table} about the survey on PFL methods, we can see that several remarks need to be considered for future directions, as follows:

\paragraph{Reliable computation with multiple models}: The existing PFL methods still rely on a centralized strategy where the server orchestrates all processes, demanding high communication and incurring computational costs. Furthermore, most methods employ a single global model, which is not suitable and poses challenges for achieving PFL due to notable dissimilarities in data distribution among clients (i.e., non-IID). To address the challenges mentioned, a solution that combines edge computing with a multi-model approach should be considered. This approach aims to reduce communication overhead, adapt models to client-specific data distributions, and enhance the overall efficiency and effectiveness of PFL.

\paragraph{Privacy-preserving}: While the PFL framework shows potential as a more favorable approach in contrast to the traditional centralized model training framework, it encounters notable difficulties, particularly concerning the selection of users for the model training procedure and the technique of augmenting data. Moreover, the problem shifts even worse when malicious clients exist and conduct various adversarial activities, such as providing incorrect information messages known as false data injection attacks [14] during collaborative training. Therefore, several privacy techniques, such as secure multiparty computation, homomorphic encryption, differential privacy, and trusted execution environment (TEE), might be leveraged for PFL in further research.

\paragraph{Trustworthy PFL}: Recent studies have shown FL flaws that may possibly endanger client security and privacy \cite{ref29}. Hence, it is of utmost importance to study attack methods on FL and formulate protective strategies to counteract them, thereby guaranteeing the security of the FL system. In order to develop resilient PFL approaches, additional research is necessary to delve into various forms of attacks and corresponding safeguards, especially as more intricate PFL protocols and structures are being introduced. Additionally, designing incentive mechanisms is a viable study area for achieving these goals to maintain fairness and motivate clients' contributions. Blockchain, as a distributed ledger technology, may be used to construct decentralized incentive systems \cite{ref30}, \cite{ref31} . These systems can play a crucial role in promoting the development of future collaborative PFL scenarios

\paragraph{Representative Datasets and Benchmark}: The growth of the PFL field depends on representative datasets. Additional datasets encompassing a wider range of modalities like sensor data, video, and audio, and encompassing a more diverse set of machine learning tasks pertinent to real-world use cases, are necessary to underpin PFL investigations. Moreover, conducting performance benchmarking holds equal importance as a pivotal factor in fostering the continuous advancement of the PFL research domain in the long run.

\section{Conclusion}
\label{sect:conclusion}
The notion of personalized federated learning is introduced to tackle statistical heterogeneity that leads to poor convergence and a lack of personalization due to the influence of non-IID data distribution among clients. This paper comprehensively summarizes the present research landscape in personalized federated learning (PFL). It also offers a concise explanation of the PFL concept and investigates associated techniques and ongoing efforts, encompassing data-based, model-based, and similarity-based methods. Lastly, this paper also discusses the need for further research and development in the field of privacy-preserving and trustworthy PFL. Additionally, it highlights the need to use representative datasets and benchmarks to ensure the effectiveness and reliability of PFL methods.

\section* {Acknowledgments}
This research was supported by the Republic of Korea’s MSIT (Ministry of Science and ICT), under the ICT Convergence Industry Innovation Technology Development Project (2022-0-00614) supervised by the IITP and partially supported by the Basic Science Research Program through the National Research Foundation of Korea (NRF) funded by the Ministry of Education (No. 2021R1I1A3046590). 



\end{document}